\documentclass[runningheads]{llncs}
\usepackage[T1]{fontenc}
\usepackage{graphicx}

\usepackage{color}
\usepackage{graphicx}
\usepackage{comment}
\usepackage{mathtools}
\usepackage{multirow}
\usepackage{placeins}
\usepackage{xcolor}
\usepackage{xspace}
\usepackage{amsmath}
\usepackage{tabularx}
\usepackage{hyperref}
\usepackage{lipsum}
\usepackage{todonotes}
\usepackage[acronym]{glossaries}
\usepackage{caption}
\usepackage{subcaption}

\usepackage{booktabs}
\usepackage{multirow}
\usepackage{lscape}

\usepackage{svg}

\usepackage[numbers,sort&compress]{natbib}

\newacronym{mKGQA}{mKGQA}{Multilingual Knowledge Graph Question Answering}
\newacronym{KGQA}{KGQA}{Knowledge Graph Question Answering}
\newacronym{QALDPlus}{QALD-9-plus}{QALD-9-Plus Dataset}
\newacronym{QALD}{QALD}{Question Answering over Linked Data}
\newacronym{QA}{QA}{Question Answering}
\newacronym{KG}{KG}{Knowledge Graph}
\newacronym{KGs}{KGs}{Knowledge Graphs}
\newacronym{RDF}{RDF}{Resource Description Framework}
\newacronym{NLP}{NLP}{Natural Language Processing}
\newacronym{NEAMT}{NEAMT}{Named Entity Aware Machine Translation}
\newacronym{NMT}{NMT}{Neural Machine Translation}
\newacronym{MT}{MT}{Machine Translation}
\newacronym{API}{API}{Application Programming Interface}
\newacronym{NEL}{NEL}{Named Entity Linking}
\newacronym{NED}{NED}{Named Entity Disambiguation}
\newacronym{NER}{NER}{Named Entity Recognition}
\newacronym{NE}{NE}{Named Entity}
\newacronym{NEs}{NEs}{Named Entities}
\newacronym{SPARQL}{SPARQL}{SPARQL}
\newacronym{QAnswer}{QAnswer}{QAnswer}
\newacronym{URI}{URI}{Uniform Resource Identifier}
\newacronym{IR}{IR}{Information Retrieval}
\newacronym{LM}{LM}{Language Model}
\newacronym{NsPM}{NsPM}{Neural SPARQL Machines}

\newcommand{\approach}{MST5\xspace}

\begin{document}
\title{MST5 --- A Transformers-Based Approach to Multilingual Question Answering over Knowledge Graphs }
\titlerunning{MST5}
\author{Nikit Srivastava \and Mengshi Ma \and Daniel Vollmers \and Hamada Zahera \and Diego Moussallem \and Axel-Cyrille Ngonga Ngomo}
\authorrunning{N. Srivastava et al.}
\institute{Universität Paderborn, Warburger Str. 100, 33098 Paderborn, Germany, \\\email{nikit.srivastava@uni-paderborn.de}}

\maketitle
\begin{abstract}
Knowledge Graph Question Answering (KGQA) simplifies querying vast amounts of knowledge stored in a graph-based model using natural language. However, the research has largely concentrated on English, putting non-English speakers at a disadvantage. Meanwhile, existing multilingual KGQA systems face challenges in achieving performance comparable to English systems, highlighting the difficulty of generating SPARQL queries from diverse languages.
In this research, we propose a simplified approach to enhance multilingual KGQA systems by incorporating linguistic context and entity information directly into the processing pipeline of a language model. Unlike existing methods that rely on separate encoders for integrating auxiliary information, our strategy leverages a single, pretrained multilingual transformer-based language model to manage both the primary input and the auxiliary data. Our methodology significantly improves the language model's ability to accurately convert a natural language query into a relevant SPARQL query.
It demonstrates promising results on the most recent QALD datasets, namely QALD-9-Plus and QALD-10. Furthermore, we introduce and evaluate our approach on Chinese and Japanese, thereby expanding the language diversity of the existing datasets.

\keywords{multilingual knowledge graphs question answering \and multilingual language models \and multilingual question answering \and sparql generation \and kgqa \and kbqa}
\end{abstract}
\section{Introduction}\label{sec:introduction}

The aim of research in Knowledge Graph Question Answering (\acrshort{KGQA}) is to establish an interactive methodology enabling users to access extensive knowledge stored within a graph-based model via natural language queries \cite{Diefenbach2018CoreTechniques}. Recent research efforts have witnessed a notable upsurge in addressing \acrshort{KGQA} concerns~\cite{qald10}. However, it is noteworthy that a substantial proportion of these systems is confined to the English language domain \cite{qanary, ganswer, soru-marx-nampi2018, sparql-qa, tebaqa, sgpt}.  Furthermore, among the currently available multilingual systems \cite{deeppavlov, platypus, qanswer}, only a limited subset of widely spoken or written languages are supported~\cite{perevalov2022can, lfqa}. Moreover, majority of these systems do not achieve performance levels comparable to those attained in English when addressing questions in other languages. This difference highlights the inherent difficulty \acrshort{KGQA} systems face in processing the multilingual natural language queries by trying to recognize the hidden, repeated patterns that are common across different languages. This poses a challenge for the vast majority of web users whose native language is not English.

\glspl{KG} have been conceptualized as language-agnostic solution for the organization and storage of information \cite{semantic-web, info13040161}. As a result, the approach to \acrshort{KGQA} paves the way for retrieving language-agnostic answers, contingent upon its ability to comprehend natural language questions in a multilingual context. Recent advancements in the field of language modeling have made numerous state-of-the-art language models freely accessible \cite{xlnet, MT5, gpt-neo, bloom}. These language models can be effectively employed within a sequence-to-sequence task setting to tackle \acrshort{KGQA}. In this setting, the input sequence corresponds to the natural language question, while the output entails a relevant \acrshort{SPARQL}\footnote{\url{https://www.w3.org/TR/rdf-sparql-query/}} query required to extract the answer from an underlying \acrshort{KG}. This method of carrying out \acrshort{KGQA} is also known as Semantic Parsing~\cite{10.1007/978-981-15-1956-7_8}, it generates an intermediate representation (i.e. \acrshort{SPARQL}) that is interpretable by humans, enabling them to understand how the model formulates specific answers within a multilingual context. 

To overcome the limited availability of multilingual \acrshort{KGQA} and the shortcomings of existing systems, we adopt a strategy similar to \citet{sgpt} that entails the inclusion of linguistic context as an auxiliary input to a language model. Unlike the previous methods, however, we employ a single encoder-decoder transformer model rather than creating separate encodings for the auxiliary input. We do this in order to leverage the attention mechanism embedded within the language model to facilitate the creation of an implicit understanding and representation of the provided linguistic context. Furthermore, we apply entity disambiguation tools to extract pertinent entity information, which is then also added to the auxiliary input. Our method establishes a seamless end-to-end system that can respond to multilingual queries using only the text of the input question.

Our findings demonstrate that incorporating linguistic context and entity information significantly enhances the \acrshort{KGQA} performance. Our approach yields promising results on our benchmarking datasets. Moreover, to substantiate the efficacy of our approach, we introduce and evaluate its performance in previously unrepresented languages within the QALD datasets. Specifically, we added Japanese to the QALD-10 dataset and both Chinese and Japanese to the QALD-9-Plus dataset. The inclusion of these non-European languages introduces a novel dimension to the datasets, given their distinctive structural characteristics when compared to the existing language set. Additionally, we make our source code publicly accessible for the benefit of the research community \footnote{\url{https://github.com/dice-group/MST5}}.
\clearpage
In essence, our approach addresses the following research question: \textit{How can the simplified integration of linguistic context and entity information into language models enhance their performance on multilingual Knowledge Graph Question Answering tasks?}  \\ 

\section{Related Work}\label{sec:related-work}
The \acrshort{KGQA} task is commonly solved by Semantic Parsing approaches, that create logical queries for given natural language questions~\cite{10.1007/978-981-15-1956-7_8}. Since \acrshort{SPARQL} is used as a standard query language for the RDF-based Knowledge Graphs~\footnote{\url{https://www.w3.org/RDF/}}, we investigate the systems that adopt this particular approach to KGQA. In this section, we cover the relevant previous works that have developed systems based on \acrfull{LM} for the \acrshort{SPARQL} query generation and the \acrshort{KGQA} systems that are multilingual.

\subsection{LM-based SPARQL query generation}

As one of the early works, \citet{soru-marx-nampi2018} introduced \acrfull{NsPM} where \acrshort{SPARQL} is regarded as an analogous language, it utilized \acrfull{NMT} to transform natural language questions into \acrshort{SPARQL} queries, while avoiding language heuristics, statistics, or manual models. The architecture consists of three components: a \texttt{generator} that creates \acrshort{SPARQL} queries from query templates for training the model, a \texttt{learner} based on Bidirectional Recurrent Neural Networks (Bi-RNNs)~\cite{brnn} to learn to translate input questions to encoded \acrshort{SPARQL}, and an \texttt{interpreter} for reconstructing \acrshort{SPARQL} queries from their encoded representation and retrieving results from a knowledge graph. Their modular design allowed integration of various machine translation models and generating \acrshort{SPARQL} queries for diverse knowledge graphs.

In the recent years, \citet{sparql-qa} produced a similar AI system for \acrshort{KGQA}, where the architecture uses a \acrshort{NMT} module based on Bi-RNNs~\cite{brnn}. They trained it in parallel to a \acrfull{NER} module, implemented using a BiLSTM-CRF network \cite{blstmcrf}. The \acrshort{NMT} module translates the input natural language question into a \acrshort{SPARQL} query template, whereas the \acrshort{NER} module extracts the entities from the question. The two modules' outputs are combined to form the resulting \acrshort{SPARQL} query. Departing from the template oriented methodology,
\citet{sgpt} proposed a novel approach to tackle the challenges associated with generating \acrshort{SPARQL} queries from natural language questions. The authors introduced a new embedding technique that encodes questions, linguistic features, and optional knowledge to allow the system to comprehend complex question patterns and graph data for \acrshort{SPARQL} query generation. They augmented the input embedding with the extracted embeddings and then fed it to a decoder-only (GPT-2~\cite{gpt2}) language model to generate \acrshort{SPARQL} query. However, it is important to highlight that the investigation conducted in the study did not assess the effectiveness of the resulting \acrshort{SPARQL} queries in retrieving answers from a \acrshort{KG}. This limitation stems from the fact that the generated \acrshort{SPARQL} queries are often syntactically incorrect. Furthermore, certain other limitations are observed within this approach, particularly regarding the selection of the language model architecture and the utilization of separate embedding layers for encoding the linguistic context.

\subsection{Multilingual KGQA}

When it comes to the multilingual \acrshort{KGQA} systems, \citet{deeppavlov} introduced a conversational system called DeepPavlov that operates by utilizing a suite of extensive language-dependent deep learning models. These models are used to execute a spectrum of tasks encompassing query type prediction, entity recognition, relation identification, and path ranking that can be applied to address \acrshort{KGQA} alongside many other uses-cases. Developing it further, \citet{deeppavlov-2023} introduced DeepPavlov-2023, which is an improved version of its predecessor.

\citet{platypus} presented a \acrshort{KGQA} methodology named Platypus, which adopts a dual-phase strategy. Initially, it leverages a semantic parser to convert questions posed in natural language into \acrshort{SPARQL} queries. Subsequently, these \acrshort{SPARQL} queries are executed on the Wikidata knowledge base to procure answers. The semantic parser employed is of a hybrid nature, integrating grammatical (rule-based) and template-based approaches. Grammatical approaches are utilized to extract the syntactic structure of the question and to identify any mentioned entities and relations. Following this, template-based methods are employed to populate predefined templates that are tailored to various types of questions.
Building towards an end-to-end multilingual \acrshort{KGQA} solution, \citet{qanswer} introduced an approach named QAnswer that generates a \acrshort{SPARQL} in four steps. First, it fetches the relevant resources from the underlying \glspl{KG}. Then, it generates a list of possible query templates. Afterwards, the query candidates are created and ranked using the output from the previous two steps. Finally, a confidence score is computed for each of the ranked queries. QAnswer claimed to be state-of-the-art at the time, and remains a top contender in the QALD challenges. However, one of the biggest downside of this system is not being open-source.

Looking at \acrfull{MT} as a viable alternative, \citet{perevalov2022can} experimented with various \acrshort{MT} systems to extend the supported languages of existing multilingual \acrshort{KGQA} systems. Their findings indicate that, in most cases, translating questions to English resulted in the highest performance. Furthermore, they noted a small to moderate positive correlation between the quality of machine translation and the question-answering score.
In a similar fashion, \citet{lfqa} adopted an entity-aware \acrshort{MT} approach for the \acrshort{KGQA} use-case. Where they highlighted the need for effective translation for the entity labels between languages. By using an entity-aware \acrshort{MT} pipeline and translating each question to English, they created a multilingual \acrshort{KGQA} approach that performs better than the traditional \acrshort{MT}-based methods previously mentioned.

\section{\approach Approach}
\label{sec:approach}
Our approach, referred to as \approach, emphasizes on incorporating and utilizing additional knowledge, such as entity link tags and linguistic context, via a transformer-based model (i.e mT5~\cite{MT5}). This is done to construct a semantic parsing-based \acrshort{KGQA} system. 
In the following subsections, we first formalize the problem definition in mathematical terms, and then describe the architecture of our approach to address this problem.

\subsection{Problem Definition}

Given a natural language query \( Q \) and auxilliary information \( A \) which consists of linguistic context and entity information, we train the transformer-based model \( \mathcal{M} \) parameterized by \( \theta \), aiming to generate a syntactically and semantically correct corresponding \acrshort{SPARQL} query \( \hat{S} \) as:

\[
\hat{S} = \arg\max_{S'} P(S' \mid Q, A; \theta)
\]

Here \( P \) is the conditional probability distribution function and  \( S' \) represents a candidate \acrshort{SPARQL} query generated by the model. \\ \\
\noindent
The model is trained in a manner akin to mT5~\cite{MT5},  focusing on minimizing the negative log likelihood (\(\mathcal{L} \)) of the correct \acrshort{SPARQL} query given the inputs:

\[
\mathcal{L}(\theta) = -\log P(S \mid Q, A; \theta)
\]

Here \( S \) refers to the reference (correct) \acrshort{SPARQL} query.

\subsection{\approach Architecture}

To obtain the auxiliary information from a given input, we create individual modules to perform entity recognition and disambiguation along with linguistic feature extraction. The extracted information is then used to compose the final input sequence to our language model by concatenating it with the input question. Figure~\ref{fig:approach-overview} provides a high-level overview of our approach. 
Following are the detailed description of these steps:

\begin{figure*}[t!]
    \centering
    \includegraphics[width=\linewidth]{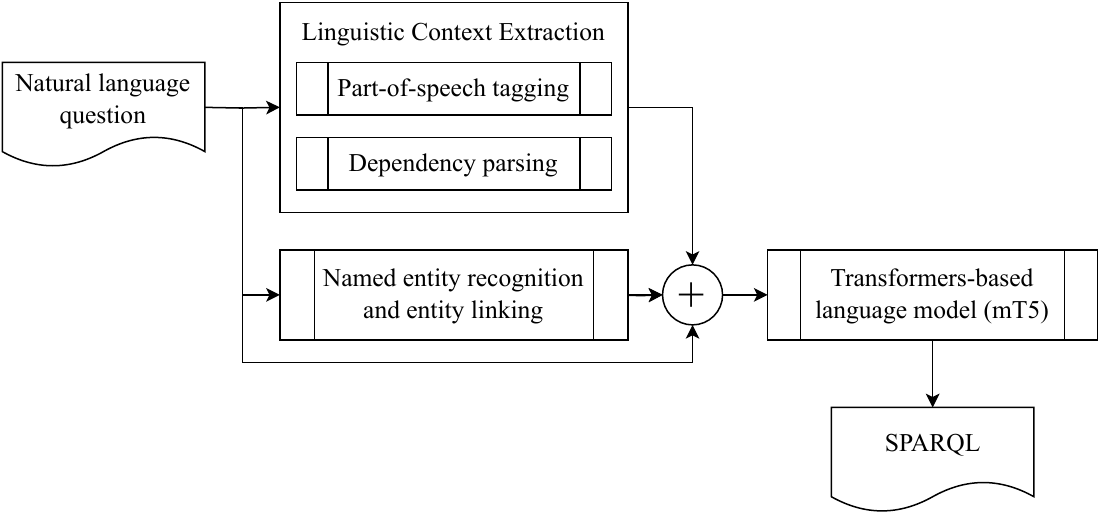}
    \caption{An overview of the MST5 approach (from left to right). First, linguistic context and entity information is extracted from the input natural language question. Then, the extracted information is concatenated with the input before being passed on to the language model. The language model generates the resulting \acrshort{SPARQL} query.}
    \label{fig:approach-overview}
\end{figure*}

\subsubsection{Named Entity Recognition and Linking}

In this step of our approach we make use of \acrshort{NER} and \acrfull{NED} approaches. To fulfill this requirement, we rely on  the \acrfull{NEAMT}\footnote{\url{https://github.com/dice-group/LFQA/tree/main/naive-eamt}} tool introduced by \citet{lfqa}. While the primary purpose of this tool is to perform entity-aware machine translation, its API allows us to skip the translation part entirely and only perform entity recognition and linking tasks. Given its multilingual capabilities, \acrshort{NEAMT} meets our requirements.

\subsubsection{Linguistic feature extraction}

For enhancing the understanding of the language model, we explicitly include linguistic features in a manner akin to the approach described by \citet{sgpt}. Unlike their approach, we directly append the features to the input sequence. To this end, we extract Part-of-Speech (POS) tags and generate the dependency tree for the input question. For extracting these features, we rely on the spaCy\footnote{\url{https://spacy.io/}} tool. For each distinct language, we use the respective spaCy model\footnote{\url{https://spacy.io/usage/models}}.

\subsubsection{Preprocessing the target SPARQL-queries}

To generate \acrshort{SPARQL} queries through a \acrshort{LM}, existing works similar to \citet{Banerjee_2022} introduce a separate vocabulary to make it easier for a \acrshort{LM} to predict the resulting \acrshort{SPARQL}. However, we recognize that this incorporation introduces additional complexity. Thus, we employ minimal preprocessing during training phase to avoid the need for an expanded vocabulary.

\acrshort{SPARQL} queries often include prefixes to shorten resource identifiers (URIs) such as \texttt{https://www.wikidata.org/entity/Q5} to \texttt{wd:Q5}. To simplify the target sequences for the mT5 model, we remove the prefix section from the query. We use the prefix form, e.g.,~\texttt{wd:Q5} instead of the full URIs in the query. During inference, we add all common prefixes back to the predicted query to restore its syntactic correctness.
As an additional step, we replace special characters such as the question mark (\texttt{?}), which signals a variable in a \acrshort{SPARQL} query, and curly braces (\texttt{\{\}}), with standardized string placeholders. This modification is due to our observation that the mT5 model struggles with accurately predicting these symbols. During the inference phase, we revert these placeholders to their original symbols to facilitate the retrieval of answers from a SPARQL endpoint.

\subsubsection{Creating Input Data for the MST5 Model}

As input to our model is a token sequence, we concatenate the textual question with the linguistic features and entity links with the help of custom separator tokens. As a result, each input sequence incorporates the features listed below. Figure~\ref{fig:linguistic-context} offers an overview of these features:

\begin{figure*}[t!]
    \centering
    \includegraphics[width=0.75\linewidth]{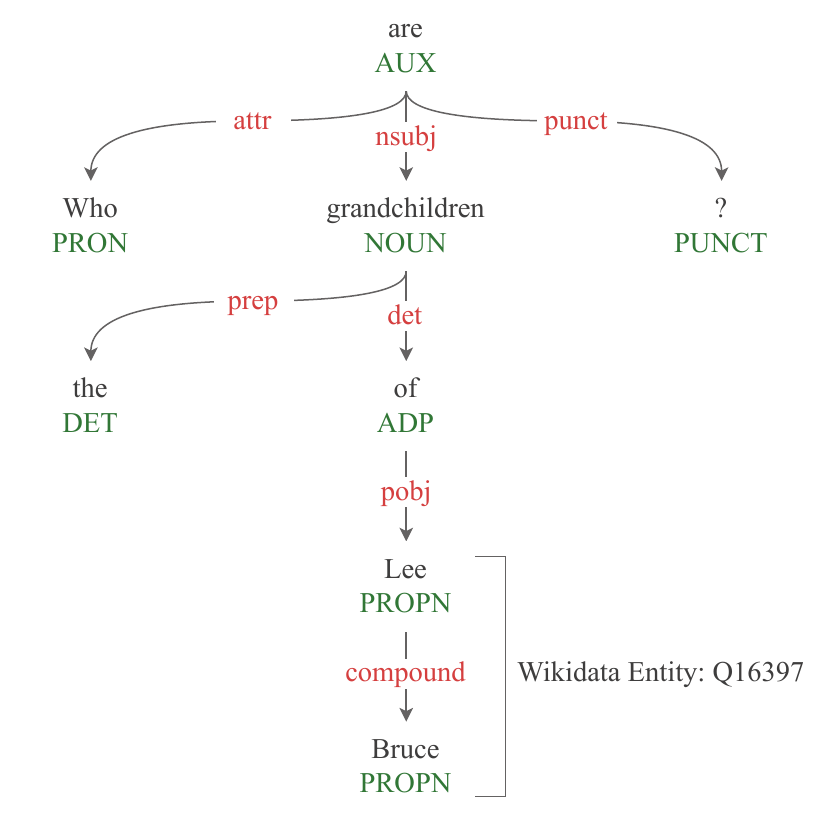}
    \caption{An example of linguistic context including dependency parsing (red) and POS-tags (green) alongisde the disambiguated entity for the text: \textit{Who are the grandchildren of Bruce Lee?}}
    \label{fig:linguistic-context}
\end{figure*}

\begin{enumerate}
    \item Question, e.g., \textit{Who are the grandchildren of Bruce Lee?}
    \item POS tags, e.g., \textit{PRON AUX DET NOUN ADP PROPN PROPN PUNCT}
    \item Dependency tree tags, e.g., \textit{attr ROOT det nsubj prep compound pobj punct}
    \item Depths in the dependency tree, e.g., \textit{2 1 3 2 3 5 4 2}
    \item Entity links, e.g., \textit{Q16397}
\end{enumerate}

Padding tokens are appended to each input feature, ensuring that the tokens for every input feature begin at a consistent token index within the tokenized input sequence. This step enables the transformers-based \acrshort{LM} to attend and contextualize the relation between the original query and the auxiliary features properly. Finally, we apply a tokenizer to the concatenated sequence to produce valid inputs for MST5 model. We use the SentencePiece~\cite{kudo-richardson-2018-sentencepiece} tokenizer for mT5~\cite{MT5} model as described in the model documentation \footnote{\url{https://huggingface.co/docs/transformers/model_doc/mt5}}. 

\section{Experimental Setup}\label{sec:experiments}

This section provides a comprehensive description of the training and benchmarking datasets used in our experiments, along with an in-depth explanation of our model training methodologies. We have organized our model training into two main categories: Ablation Study and Evaluation of Optimal Models. This structure helps with systematic investigation of the effects of various model configurations and enables us to concentrate on the best-performing models for additional analysis. Our current experimental setup is aimed at addressing the following research questions:
\begin{enumerate}
    \item \textit{Whether the simplified auxilliary input of linguistic context and entity information has any effect on the KGQA performance?}
    \item \textit{How the top-performing ablations compare to other systems?}
    \item \textit{How the the optimal ablations perform in a multilingual setting?}
\end{enumerate}

\subsection{Training and Evaluation Datasets}
To train the \approach model, we utilize the LC-QuAD 2.0, QALD-9-Plus, and QALD-10 datasets. Below, we outline these datasets and detail any modifications applied:
\\ \\
\textbf{LC-QuAD 2.0}: Introduced by \citet{lcquad2}, is the largest English-only \acrshort{KGQA} dataset. It consists of over 30,000 natural language questions paired with their corresponding Wikidata SPARQL queries. We choose this dataset primarily to train the model on contextualizing SPARQL queries alongside natural language queries before proceeding to train on a smaller, multilingual dataset. \\

\noindent
\textbf{QALD-9-Plus}: Introduced by \citet{perevalov2022qald}, QALD-9-Plus dataset includes the questions in English, German, Russian, French, Spanish, Armenian, Belarusian, Lithuanian, Bashkir and Ukrainian.  The training set for Wikidata consists of 371 questions, while the test set has 136 questions. We update the QALD-9-Plus dataset by adding Chinese and Japanese translations of each question. These translations were done by a native speaker of these languages. Since the test set contains the answers from an older (unavailable) version of Wikidata, we update the QALD-9-Plus dataset using endpoint provided for QALD-10\footnote{\url{https://github.com/KGQA/QALD-10\#endpoint}} as both of these datasets were created concurrently. Subsequently, we filter out all the questions with empty answer-set. The refined test dataset now comprises 102 questions. We refer to this version as QALD-9-Plus (updated) test dataset in the upcoming sections. \\

\noindent
\textbf{QALD-10}: Introduced by \citet{qald10}, it comprises of questions in English, German, Chinese and Russian. It uses QALD-9-Plus as its training data, and provides a test dataset of 394 questions. We update the QALD-10 dataset by adding Japanese translation for each question. These translations were done by a native speaker of the respective language. 
\\ \\
\noindent
To benchmark our models, our approach requires datasets that are multilingual and provide a reference \acrshort{SPARQL} query alongside the retrieved answer-set for the respective natural language question. These criteria are essential to show that our models can generate \acrshort{SPARQL} queries that are not only similar to the reference but also possess the capability to fetch the accurate answer. To fulfil these requirements, we make use of QALD-9-Plus (updated) test and QALD-10 dataset for evaluation.
A key distinction among these QALD datasets lies in the evaluation metric employed for comparison. The QALD-9-Plus dataset utilizes \texttt{Macro F1} for assessing the performance across various \acrshort{KGQA} systems, whereas the QALD10 dataset suggests \texttt{Macro F1 QALD}\footnote{\url{https://github.com/dice-group/gerbil/issues/320}} as per community's request\footnote{\url{https://github.com/dice-group/gerbil/issues/211}}. To conduct the evaluation and obtain performance metrics on these datasets, we make use of the GERBIL-QA~\cite{gerbil-qa} tool.

\begin{table*}[]
\centering
\begin{tabular}{@{}lllll@{}}
\toprule
\textbf{Name}                               & \textbf{Ling. Ctxt.} & \textbf{Ent. Info.} & \textbf{LC-QuAD 2.0} & \textbf{QALD-9-Plus} \\ \midrule
-                                           & No                     & No                   & No                         & No                           \\
MST5$_{\text{Q9}}$        & No                     & No                   & No                         & Yes                          \\
MST5$_{\text{L2}}$          & No                     & No                   & Yes                        & No                           \\
MST5$_{\text{L2+Q9}}$ & No                     & No                   & Yes                        & Yes                          \\
-                                           & No                     & Yes                  & No                         & No                           \\
MST5$_{\text{Q9+ENT}}$           & No                     & Yes                  & No                         & Yes                          \\
MST5$_{\text{L2+ENT}}$             & No                     & Yes                  & Yes                        & No                           \\
MST5$_{\text{L2+Q9+ENT}}$    & No                     & Yes                  & Yes                        & Yes                          \\
-                                           & Yes                    & No                   & No                         & No                           \\
MST5$_{\text{Q9+LING}}$            & Yes                    & No                   & No                         & Yes                          \\
MST5$_{\text{L2+LING}}$              & Yes                    & No                   & Yes                        & No                           \\
MST5$_{\text{L2+Q9+LING}}$     & Yes                    & No                   & Yes                        & Yes                          \\
-                                           & Yes                    & Yes                  & No                         & No                           \\
MST5$_{\text{Q9+LING+ENT}}$        & Yes                    & Yes                  & No                         & Yes                          \\
MST5$_{\text{L2+LING+ENT}}$           & Yes                    & Yes                  & Yes                        & No                           \\
MST5$_{\text{L2+Q9+LING+ENT}}$ & Yes                    & Yes                  & Yes                        & Yes                          \\ \bottomrule
\end{tabular}
\vspace{.5em}
\caption{All covered MST5 ablations based on the auxiliary features (linguistic context and entity information) and fine-tuning dataset used during the model training.}
\label{tab:mst5-ablation-table}
\end{table*}

\subsection{Model Training Details}
We selected the pretrained mT5\footnote{\url{https://huggingface.co/google/mt5-xl}} as the foundational model for our work. Its pretraining involved the mC4\footnote{\url{https://www.tensorflow.org/datasets/catalog/c4\#c4multilingual}} corpus, encompassing 101 diverse languages.

During the fine-tuning phase, we set the maximum number of epochs to 300 while setting up an early-stop regularization method to prevent overfitting. Also, we use the DeepSpeed\footnote{\url{https://www.deepspeed.ai/}} tool introduced by \citet{rajbhandari2020zero} to optimize the computing resource usage.
As for the hardware, we make use of the Nvidia-A100\footnote{\url{https://www.nvidia.com/en-us/data-center/a100/}} GPU for training our models.
\subsection{Ablation Study} 

To assess the impact of the proposed input augmentations on the end-to-end \acrshort{KGQA} performance of our approach, we conduct an ablation study. Our study involves training and testing every potential variant that includes the following modifications:
\begin{itemize}
    \item Fine-tuning on LC-QuAD 2.0 dataset
    \item Fine-tuning on QALD-9-Plus dataset
    \item Adding linguistic context to the input
    \item Incorporating entity tags into the input
\end{itemize}

Table \ref{tab:mst5-ablation-table} lists all the covered ablations. We evaluate these ablations on the English questions from QALD-9-Plus (updated) test dataset to compare their performance. We exclude the multilingual questions due to resource limitations, as for each additional language all the ablations would need to be evaluated. We also exclude QALD-10 dataset from the ablation study because its training data is a combination of the QALD-9-Plus training and testing data.

\subsection{Optimal Models Evaluation}

From the ablation study, we select the model(s) demonstrating the best performance and subject them to further evaluation, during this evaluation we cover all the supported languages.

For QALD-9-Plus (updated) test dataset, we compare the performance of our optimal models with only one other \acrshort{KGQA} system: DeepPavlov-2023~\citep{deeppavlov-2023}. We were not able to include any other relevant system \cite{qanswer, platypus, deeppavlov} as we couldn't extract the generated \acrshort{SPARQL} queries for updated QALD-9-Plus dataset\footnote{Either unreachable public APIs, technical issues in the local deployment or no SPARQL query generation.}. 
For QALD-10, we refine our optimal model(s) by additional training on a merged dataset of QALD-9-Plus train and test data. The further fine-tuning allowed us to perform a direct comparison with the results of the other existing models in a fair manner.
We made use of the QA-System-Wrapper\footnote{\url{https://github.com/WSE-research/qa-systems-wrapper}} tool to query the DeepPavlov-2023 system.

\begin{table*}[]
\centering
\begin{tabular}{@{}lrrrr@{}}
\toprule
\textbf{Ablation name}                      & \multicolumn{1}{l}{\textbf{F1}} & \multicolumn{1}{l}{\textbf{Precision}} & \multicolumn{1}{l}{\textbf{Recall}} & \multicolumn{1}{l}{\textbf{F1 QALD}} \\ \midrule
MST5$_{\text{L2+ENT}}$             & 0.1886                                & 0.2001                                       & 0.1982                                    & 0.322                                      \\
MST5$_{\text{L2+LING}}$              & 0.2291                                & 0.2355                                       & 0.2353                                    & 0.367                                      \\
MST5$_{\text{L2+LING+ENT}}$           & 0.2622                                & 0.2747                                       & 0.2698                                    & 0.4088                                     \\
MST5$_{\text{L2}}$          & 0.1353                                & 0.134                                        & 0.1373                                    & 0.2359                                     \\
\midrule
MST5$_{\text{Q9+ENT}}$           & 0.3173                                & 0.3318                                       & 0.3164                                    & 0.4554                                     \\
MST5$_{\text{Q9+LING}}$            & 0.206                                 & 0.2188                                       & 0.2043                                    & 0.2759                                     \\
MST5$_{\text{Q9+LING+ENT}}$        & 0.3203                                & 0.3424                                       & 0.3216                                    & 0.4624                                     \\
MST5$_{\text{Q9}}$        & 0.0098                                & 0.0098                                       & 0.0098                                    & 0.0194                                     \\
\midrule
MST5$_{\text{L2+Q9+ENT}}$    & \textbf{0.4187}                       & \textbf{0.4366}                              & \textbf{0.4242}                           & \textbf{0.572}                             \\
MST5$_{\text{L2+Q9+LING}}$     & 0.3407                                & 0.3571                                       & 0.3493                                    & 0.4745                                     \\
MST5$_{\text{L2+Q9+LING+ENT}}$ & 0.4015                                & 0.4192                                       & 0.4046                                    & 0.5563                                     \\
MST5$_{\text{L2+Q9}}$ & 0.2628                                & 0.2792                                       & 0.2631                                    & 0.3663                                     \\
\bottomrule
\end{tabular}
\vspace{.5em}
\caption{Macro performance metrics of MST5 ablations on the QALD-9-Plus (updated) test dataset (English): divided into three groups based on fine-tuning data, separated by horizontal lines}
\label{tab:mst5-ablation-performance}
\end{table*}

\section{Results and Discussion}\label{sec:results-dicussion}

Our discussion of the results is structured into three sections: Initially, in \autoref{english-based-performance}, we focus on the performance metrics for English. Then, in \autoref{multilingual-based-performance}, we examine the performance of our best ablations across various languages. Finally, in \autoref{subsec:discussion}, we discuss the challenges encountered during our evaluation with external systems and the measures we implemented to prevent our system from contributing to the same problem.

\subsection{English-based Performance}
\label{english-based-performance}

As a first step, we perform the ablation study to find out \textit{whether the simplified auxilliary input of linguistic context and entity information has any effect on the KGQA performance}. We evaluate all our model ablations on the QALD-9-Plus (updated) English dataset. Table \ref{tab:mst5-ablation-performance} provides a direct comparison of the model ablations. Based on the results (Macro F1), we observe the following:

\begin{itemize}
    \item Linguistic context and entities always improve the performance of the system
    \item In most cases, combination of linguistic context and entities leads to the best performing model
    \item Between the models fine-tuned on both datasets, the entity-only variant has slightly better performance
\end{itemize}

Now, to investigate \textit{how the top-performing ablations compare to other systems}, we pick the two best performing models \texttt{MST5$_{\text{L2+Q9+ENT}}$}, \texttt{MST5$_{\text{L2+Q9+LING+ENT}}$} and simplify their names to \texttt{MST5$_{\text{ENT}}$}, \texttt{MST5$_{\text{LING+ENT}}$} respectively. Table~\ref{tab:qald10-comparison} presents a comparative analysis of various models tested on the QALD-10 dataset. In this comparison, we find that:

\begin{itemize}
    \item \approach significantly outperforms the competing systems
    \item Model variant incorporating both linguistic context and entity information emerges as the top performer
\end{itemize}

\begin{table}[]
\centering
\begin{tabular}{@{}lrrrr@{}}
\toprule
\textbf{Approach}          & \multicolumn{1}{l}{\textbf{F1}} & \multicolumn{1}{l}{\textbf{Precision}} & \multicolumn{1}{l}{\textbf{Recall}} & \multicolumn{1}{l}{\textbf{F1 QALD}} \\ \midrule
\citeauthor{sparql-qa} & 0.4538                                & 0.4538                                       & 0.4574                                    & 0.5947                                     \\
\citeauthor{qanswer}                    & 0.5070                                & 0.5068                                       & 0.5238                                    & 0.5776                                     \\
\citeauthor{Shivashankar2022FromGraph}           & 0.3215                                & 0.3206                                       & 0.3312                                    & 0.4909                                     \\
\citeauthor{Baramiia2022RankingApproach}            & 0.4277                                & 0.4289                                       & 0.4272                                    & 0.4281                                     \\

DeepPavlov-2023~\cite{deeppavlov-2023}            & 0.3241                                & 0.3279                                       & 0.3369                                    & 0.4518                                     \\ \midrule
MST5$_{\text{ENT}}$               & 0.4784                                & 0.4777                                       & 0.4848                                    & 0.6330                                     \\
MST5$_{\text{LING+ENT}}$            & \textbf{0.5271}                       & \textbf{0.5271}                              & \textbf{0.5317}                           & \textbf{0.6727}                            \\ \bottomrule
\end{tabular}
\vspace{0.5em}
\caption{QALD-10 macro performance comparison (English).}
\label{tab:qald10-comparison}
\end{table}

\subsection{Multilingual Performance}
\label{multilingual-based-performance}

We proceed by examining \textit{how the the optimal ablations perform in a multilingual setting}. Table~\ref{tab:mst5-qald9plus-comparison-macrof1} provides direct comparison of our models \texttt{MST5$_{\text{ENT}}$} and \texttt{MST5$_{\text{LING+ENT}}$} with DeepPavlov-2023~\citep{deeppavlov-2023} on all supported languages\footnote{The language codes used in the datasets and our experiments are as per the ISO 639-1 standard \url{https://en.wikipedia.org/wiki/List_of_ISO_639-1_codes}} or QALD-9-Plus (updated). The results show the following:

\begin{itemize}
    \item \approach variants perform better than DeepPavlov-2023 for both languages supported by it (English, Russian)
    \item When comparing the \approach variants to each other:
    \begin{itemize}
        \item Performance in German and French is quite comparable to English
        \item Performance in Spanish, Russian, Lithuanian, Ukrainian, Belarusian, and Chinese is a bit lower than English
        \item For Bashkir and Japanese the performance is very low in comparison
    \end{itemize}
\end{itemize}

For Bashkir, the low performance is due to bad quality of the entity recognition and linguistic context extraction through the third-party tools. For Japanese, the lower F1 score is because of the small size of QALD-9-Plus (updated) test set, as many resultant \acrshort{SPARQL} queries for Japanese did not work.

Finally, in the Table~\ref{tab:qald10-comparison-multilingual}, we have evaluation results for multilingual setting on QALD-10. We observe that our models not only outperform the DeepPavlov-2023, but also achieve comparable results on all supported languages. The performance on Japanese is lower in comparison to other languages, but still significantly better than the QALD-9-Plus (updated) test dataset. We attribute this improvement in performance to the larger size ($\approx$4x) of the QALD-10 test dataset.

\begin{table*}[]
\centering
\begin{tabular}{@{}lrrr@{}}
\toprule
\textbf{} & \multicolumn{1}{l}{DeepPavlov-2023~\cite{deeppavlov-2023}} & \multicolumn{1}{l}{MST5$_{\text{ENT}}$} & \multicolumn{1}{l}{MST5$_{\text{LING+ENT}}$} \\ \midrule
ba        & -                                   & \textbf{0.1842}                  & 0.1579                              \\
be        & -                                   & \textbf{0.2907}                  & 0.2807                              \\
de        & -                                   & \textbf{0.4126}                  & 0.3851                              \\
en        & 0.3716                              & \textbf{0.4187}                  & 0.4015                              \\
es        & -                                   & \textbf{0.3600}                    & 0.3498                              \\
fr        & -                                   & \textbf{0.4167}                  & \textbf{0.4167}                     \\
ja        & -                                   & 0.0654                           & \textbf{0.0752}                     \\
lt        & -                                   & \textbf{0.3115}                  & 0.2792                              \\
ru        & 0.3117                              & \textbf{0.3761}                  & 0.3467                              \\
uk        & -                                   & \textbf{0.3467}                  & 0.3369                              \\
zh        & -                                   & \textbf{0.3344}                  & 0.3115                              \\ \bottomrule
\end{tabular}
\vspace{.5em}
\caption{Performance (Macro F1) comparison of MST5 with DeepPavlov-2023 on the QALD-9-Plus (updated) test dataset on all supported languages}
\label{tab:mst5-qald9plus-comparison-macrof1}
\end{table*}

\subsection{Discussion}
\label{subsec:discussion}

Keeping overall results into perspective, we observe that some of our comparison tables lack the performance data for other \acrshort{KGQA} systems. We mainly face this issue with our QALD-9-Plus (updated) dataset. For the QALD-10 dataset, the issue is with the lack of performance data on non-English languages. \\ \\
\noindent
This shortfall stems from several contributing factors:
\begin{enumerate}
    \item \acrshort{KGQA} systems~\cite{ganswer, deeppavlov} not providing a SPARQL, making it hard to run on a custom dataset and \acrshort{SPARQL} endpoint;
    \item Unresponsive endpoints~\cite{qanswer, platypus} or source code that doesn't work; 
    \item No maintained knowledge-base endpoint for the evaluation dataset~\cite{perevalov2022qald};
    \item Absence of multilingual performance numbers for the evaluated systems~\cite{qald10}.
\end{enumerate}
Due to these factors, reproducing or generating new results on existing multilingual systems becomes a challenging task. To mitigate this, it is beneficial to have open-source and well-documented systems, complemented by datasets that offer a maintained \acrshort{SPARQL} endpoint or the relevant knowledge-base data dumps.

In our approach, we offer an open-source and thoroughly documented codebase. Additionally, our experimental framework, as detailed in Section \ref{sec:experiments}, enables the replication and straightforward comparison of results, serving as a valuable reference for future research work.

\begin{table*}[t!]
\centering
\begin{tabular}{@{}lrrr@{}}
\toprule
   & \multicolumn{1}{l}{DeepPavlov-2023~\cite{deeppavlov-2023}} & \multicolumn{1}{l}{MST5$_{\text{ENT}}$} & \multicolumn{1}{l}{MST5$_{\text{LING+ENT}}$} \\ \midrule
de & -                                   & 0.5908                           & \textbf{0.6048}                     \\
en & 0.5092                              & 0.6330                            & \textbf{0.6727}                     \\
ja & -                                   & 0.4224                           & \textbf{0.4964}                     \\
ru & 0.4118                              & 0.6296                           & \textbf{0.6600}                       \\
zh & -                                   & 0.5877                           & \textbf{0.6398}                     \\ \bottomrule
\end{tabular}
\vspace{0.5em}
\caption{QALD-10 performance (Macro F1 QALD) comparison for all supported languages.}
\label{tab:qald10-comparison-multilingual}
\end{table*}

\section{Limitations}
\label{sec:limitations}

The one of the biggest limitation of \approach roots from its reliance on the third-party tools. These tools are used for entity recognition, entity disambiguation and linguistic context extraction tasks and can differ depending upon the language of the input. This can lead to issues in the future when more languages are introduced to our system, making it harder to manage a large number of third-party modules. Furthermore, in its current state, our system does not adequately support or perform well with low-resource languages, such as Armenian and Bashkir, posing a significant obstacle to achieving a genuinely multilingual system.

To address these limitations, we plan to extend our work by incorporating a multi-task approach. A multi-task approach will utilize a single \acrshort{LM} to extract linguistic context and perform entity recognition \& disambiguation. This not only solves the problem of reliance on third-party tools, but also helps create a system that can perform relatively well on low-resource languages through generalization on the related ones.
As part of future research, we also plan to extend the auxiliary input by including additional information such as entity types and relations. Furthermore, we plan to investigate if we can introduce \acrshort{NLP} techniques such as Semantic Role Labeling~\cite{JurafskyMartin2024} to further improve the quality of resulting \acrshort{SPARQL} query.

\section{Conclusion}\label{sec:conclusion}

In this work, we presented a new strategy for multilingual \acrshort{KGQA}. We built upon an existing multilingual \acrshort{LM} mT5~\cite{MT5} and combined it with an approach inspired by \citet{sgpt}. Unlike the previous work, our approach simplifies the input augmentation by making the \acrshort{LM} autonomously learn and contextualize the new representations. We performed an extensive ablation study of our model and showed how the auxiliary input significantly improves the end-to-end \acrshort{KGQA} performance on the latest QALD datasets. Furthermore, we conducted comparisons with other \acrshort{KGQA} systems, demonstrating the superior performance of our system. However, we also noted the small number of \acrshort{KGQA} systems and the general lack of data on multilingual performance, which made a comprehensive comparison challenging.
We pointed out the common issues leading to this problem and addressed them in our approach to make it easier for future research in the same direction. Lastly, we acknowledged the limitations of our method and propose a multi-task strategy, setting up the foundation for future research.

\clearpage

\bibliographystyle{unsrtnat}
\bibliography{mybib}

\begin{thebibliography}{33}
\providecommand{\natexlab}[1]{#1}
\providecommand{\url}[1]{\texttt{#1}}
\expandafter\ifx\csname urlstyle\endcsname\relax
  \providecommand{\doi}[1]{doi: #1}\else
  \providecommand{\doi}{doi: \begingroup \urlstyle{rm}\Url}\fi

\bibitem[Diefenbach et~al.(2018)Diefenbach, Lopez, Singh, and Maret]{Diefenbach2018CoreTechniques}
Dennis Diefenbach, Vanessa Lopez, Kamal Singh, and Pierre Maret.
\newblock Core techniques of question answering systems over knowledge bases: a survey.
\newblock \emph{Knowledge and Information Systems}, 55:\penalty0 529--569, 2018.
\newblock \doi{10.1007/s10115-017-1100-y}.
\newblock URL \url{https://doi.org/10.1007/s10115-017-1100-y}.

\bibitem[Usbeck et~al.(2023)Usbeck, Yan, Perevalov, Jiang, Schulz, Kraft, M{\"o}ller, Huang, Reineke, Ngomo, Saleem, and Both]{qald10}
Ricardo Usbeck, Xi~Yan, Aleksandr Perevalov, Longquan Jiang, Julius Schulz, Angelie Kraft, Cedric M{\"o}ller, Junbo Huang, Jan Reineke, Axel-Cyrille~Ngonga Ngomo, Muhammad Saleem, and Andreas Both.
\newblock {QALD-10 — The 10th Challenge on Question Answering over Linked Data}.
\newblock \emph{Under review in the Semantic Web Journal}, 02 2023.
\newblock URL \url{https://www.semantic-web-journal.net/system/files/swj3357.pdf}.

\bibitem[Both et~al.(2016)Both, Diefenbach, Singh, Shekarpour, Cherix, and Lange]{qanary}
Andreas Both, Dennis Diefenbach, Kuldeep Singh, Saeedeh Shekarpour, Didier Cherix, and Christoph Lange.
\newblock Qanary – a methodology for vocabulary-driven open question answering systems.
\newblock 05 2016.
\newblock ISBN 978-3-319-34128-6.
\newblock \doi{10.1007/978-3-319-34129-3_38}.

\bibitem[Hu et~al.(2018)Hu, Zou, Yu, Wang, and Zhao]{ganswer}
Sen Hu, Lei Zou, Jeffrey~Xu Yu, Haixun Wang, and Dongyan Zhao.
\newblock Answering natural language questions by subgraph matching over knowledge graphs.
\newblock \emph{IEEE Transactions on Knowledge and Data Engineering}, 30:\penalty0 824--837, 2018.
\newblock URL \url{https://api.semanticscholar.org/CorpusID:4569766}.

\bibitem[Soru et~al.(2018)Soru, Marx, Valdestilhas, Esteves, Moussallem, and Publio]{soru-marx-nampi2018}
Tommaso Soru, Edgard Marx, Andr\'e Valdestilhas, Diego Esteves, Diego Moussallem, and Gustavo Publio.
\newblock Neural machine translation for query construction and composition.
\newblock 2018.
\newblock URL \url{https://arxiv.org/abs/1806.10478}.

\bibitem[Borroto et~al.(2021)Borroto, Ricca, and Cuteri]{sparql-qa}
Manuel~Alejandro Borroto, Francesco Ricca, and Bernardo Cuteri.
\newblock A system for translating natural language questions into sparql queries with neural networks: Preliminary results \(discussion paper\).
\newblock In \emph{SEBD 2021: Italian Symposium on Advanced Database Systems}, pages 226--234, Aachen, Germany, 2021. RWTH Aachen.
\newblock URL \url{https://www.tib.eu/de/suchen/id/TIBKAT%3A181641011X}.

\bibitem[Vollmers et~al.(2021)Vollmers, Jalota, Moussallem, Topiwala, Ngomo, and Usbeck]{tebaqa}
Daniel Vollmers, Rricha Jalota, Diego Moussallem, Hardik Topiwala, Axel-Cyrille~Ngonga Ngomo, and Ricardo Usbeck.
\newblock Knowledge graph question answering using graph-pattern isomorphism.
\newblock In \emph{Studies on the Semantic Web}. {IOS} Press, aug 2021.
\newblock \doi{10.3233/ssw210038}.
\newblock URL \url{https://doi.org/10.3233%2Fssw210038}.

\bibitem[Rony et~al.(2022)Rony, Kumar, Teucher, Kovriguina, and Lehmann]{sgpt}
Md~Rashad Al~Hasan Rony, Uttam Kumar, Roman Teucher, Liubov Kovriguina, and Jens Lehmann.
\newblock Sgpt: A generative approach for sparql query generation from natural language questions.
\newblock \emph{IEEE Access}, 10:\penalty0 70712--70723, 2022.
\newblock \doi{10.1109/ACCESS.2022.3188714}.

\bibitem[Burtsev et~al.(2018)Burtsev, Seliverstov, Airapetyan, Arkhipov, Baymurzina, Bushkov, Gureenkova, Khakhulin, Kuratov, Kuznetsov, Litinsky, Logacheva, Lymar, Malykh, Petrov, Polulyakh, Pugachev, Sorokin, Vikhreva, and Zaynutdinov]{deeppavlov}
Mikhail Burtsev, Alexander Seliverstov, Rafael Airapetyan, Mikhail Arkhipov, Dilyara Baymurzina, Nickolay Bushkov, Olga Gureenkova, Taras Khakhulin, Yuri Kuratov, Denis Kuznetsov, Alexey Litinsky, Varvara Logacheva, Alexey Lymar, Valentin Malykh, Maxim Petrov, Vadim Polulyakh, Leonid Pugachev, Alexey Sorokin, Maria Vikhreva, and Marat Zaynutdinov.
\newblock {D}eep{P}avlov: Open-source library for dialogue systems.
\newblock In \emph{Proceedings of {ACL} 2018, System Demonstrations}, pages 122--127, Melbourne, Australia, July 2018. Association for Computational Linguistics.
\newblock \doi{10.18653/v1/P18-4021}.
\newblock URL \url{https://aclanthology.org/P18-4021}.

\bibitem[Pellissier~Tanon et~al.(2018)Pellissier~Tanon, de~Assun{\c{c}}{\~a}o, Caron, and Suchanek]{platypus}
Thomas Pellissier~Tanon, Marcos~Dias de~Assun{\c{c}}{\~a}o, Eddy Caron, and Fabian~M. Suchanek.
\newblock Demoing platypus -- a multilingual question answering platform for wikidata.
\newblock In Aldo Gangemi, Anna~Lisa Gentile, Andrea~Giovanni Nuzzolese, Sebastian Rudolph, Maria Maleshkova, Heiko Paulheim, Jeff~Z. Pan, and Mehwish Alam, editors, \emph{The Semantic Web: ESWC 2018 Satellite Events}, pages 111--116, Cham, 2018. Springer International Publishing.
\newblock ISBN 978-3-319-98192-5.

\bibitem[Diefenbach et~al.(2020)Diefenbach, Both, Singh, and Maret]{qanswer}
Dennis Diefenbach, A.~Both, K.~Singh, and P.~Maret.
\newblock Towards a question answering system over the semantic web.
\newblock \emph{Semantic Web}, 11:\penalty0 421--439, 2020.
\newblock \doi{10.3233/SW-190343}.

\bibitem[Perevalov et~al.(2022{\natexlab{a}})Perevalov, Both, Diefenbach, and Ngonga~Ngomo]{perevalov2022can}
Aleksandr Perevalov, Andreas Both, Dennis Diefenbach, and Axel-Cyrille Ngonga~Ngomo.
\newblock Can machine translation be a reasonable alternative for multilingual question answering systems over knowledge graphs?
\newblock In \emph{Proceedings of the ACM Web Conference 2022}, pages 977--986, 2022{\natexlab{a}}.

\bibitem[Srivastava et~al.(2023)Srivastava, Perevalov, Kuchelev, Moussallem, Ngonga~Ngomo, and Both]{lfqa}
Nikit Srivastava, Aleksandr Perevalov, Denis Kuchelev, Diego Moussallem, Axel-Cyrille Ngonga~Ngomo, and Andreas Both.
\newblock Lingua franca – entity-aware machine translation approach for question answering over knowledge graphs.
\newblock In \emph{Proceedings of the 12th Knowledge Capture Conference 2023}, K-CAP '23, page 122–130, New York, NY, USA, 2023. Association for Computing Machinery.
\newblock ISBN 9798400701412.
\newblock \doi{10.1145/3587259.3627567}.
\newblock URL \url{https://doi.org/10.1145/3587259.3627567}.

\bibitem[Berners-Lee et~al.(2001)Berners-Lee, Hendler, and Lassila]{semantic-web}
Tim Berners-Lee, James Hendler, and Ora Lassila.
\newblock The semantic web: A new form of web content that is meaningful to computers will unleash a revolution of new possibilities.
\newblock \emph{ScientificAmerican.com}, 05 2001.

\bibitem[Kejriwal(2022)]{info13040161}
Mayank Kejriwal.
\newblock Knowledge graphs: A practical review of the research landscape.
\newblock \emph{Information}, 13\penalty0 (4), 2022.
\newblock ISSN 2078-2489.
\newblock \doi{10.3390/info13040161}.
\newblock URL \url{https://www.mdpi.com/2078-2489/13/4/161}.

\bibitem[Yang et~al.(2020)Yang, Dai, Yang, Carbonell, Salakhutdinov, and Le]{xlnet}
Zhilin Yang, Zihang Dai, Yiming Yang, Jaime Carbonell, Ruslan Salakhutdinov, and Quoc~V. Le.
\newblock Xlnet: Generalized autoregressive pretraining for language understanding, 2020.

\bibitem[Xue et~al.(2020)Xue, Constant, Roberts, Kale, Al{-}Rfou, Siddhant, Barua, and Raffel]{MT5}
Linting Xue, Noah Constant, Adam Roberts, Mihir Kale, Rami Al{-}Rfou, Aditya Siddhant, Aditya Barua, and Colin Raffel.
\newblock mt5: {A} massively multilingual pre-trained text-to-text transformer.
\newblock \emph{CoRR}, abs/2010.11934, 2020.
\newblock URL \url{https://arxiv.org/abs/2010.11934}.

\bibitem[Black et~al.(2021)Black, Gao, Wang, Leahy, and Biderman]{gpt-neo}
Sid Black, Leo Gao, Phil Wang, Connor Leahy, and Stella Biderman.
\newblock {GPT-Neo: Large Scale Autoregressive Language Modeling with Mesh-Tensorflow}, March 2021.
\newblock URL \url{https://doi.org/10.5281/zenodo.5297715}.
\newblock {If you use this software, please cite it using these metadata.}

\bibitem[Scao et~al.(2023)Scao, Fan, Akiki, Pavlick, Ilić, and et~collab]{bloom}
Teven~Le Scao, Angela Fan, Christopher Akiki, Ellie Pavlick, Suzana Ilić, and et~collab.
\newblock Bloom: A 176b-parameter open-access multilingual language model, 2023.

\bibitem[Wu et~al.(2019)Wu, Zhang, and Feng]{10.1007/978-981-15-1956-7_8}
Peiyun Wu, Xiaowang Zhang, and Zhiyong Feng.
\newblock A survey of question answering over knowledge base.
\newblock In \emph{Knowledge Graph and Semantic Computing: Knowledge Computing and Language Understanding}, pages 86--97, Singapore, 2019. Springer Singapore.
\newblock ISBN 978-981-15-1956-7.

\bibitem[Schuster and Paliwal(1997)]{brnn}
M.~Schuster and K.K. Paliwal.
\newblock Bidirectional recurrent neural networks.
\newblock \emph{IEEE Transactions on Signal Processing}, 45\penalty0 (11):\penalty0 2673--2681, 1997.
\newblock \doi{10.1109/78.650093}.

\bibitem[Huang et~al.(2015)Huang, Xu, and Yu]{blstmcrf}
Zhiheng Huang, Wei Xu, and Kai Yu.
\newblock Bidirectional lstm-crf models for sequence tagging, 2015.

\bibitem[Radford et~al.(2019)Radford, Wu, Child, Luan, Amodei, and Sutskever]{gpt2}
Alec Radford, Jeff Wu, Rewon Child, David Luan, Dario Amodei, and Ilya Sutskever.
\newblock Language models are unsupervised multitask learners.
\newblock 2019.
\newblock URL \url{https://api.semanticscholar.org/CorpusID:160025533}.

\bibitem[Zharikova et~al.(2023)Zharikova, Kornev, Ignatov, Talimanchuk, Evseev, Petukhova, Smilga, Karpov, Shishkina, Kosenko, and Burtsev]{deeppavlov-2023}
Diliara Zharikova, Daniel Kornev, Fedor Ignatov, Maxim Talimanchuk, Dmitry Evseev, Ksenya Petukhova, Veronika Smilga, Dmitry Karpov, Yana Shishkina, Dmitry Kosenko, and Mikhail Burtsev.
\newblock {D}eep{P}avlov dream: Platform for building generative {AI} assistants.
\newblock In Danushka Bollegala, Ruihong Huang, and Alan Ritter, editors, \emph{Proceedings of the 61st Annual Meeting of the Association for Computational Linguistics (Volume 3: System Demonstrations)}, pages 599--607, Toronto, Canada, July 2023. Association for Computational Linguistics.
\newblock \doi{10.18653/v1/2023.acl-demo.58}.
\newblock URL \url{https://aclanthology.org/2023.acl-demo.58}.

\bibitem[Banerjee et~al.(2022)Banerjee, Nair, Kaur, Usbeck, and Biemann]{Banerjee_2022}
Debayan Banerjee, Pranav~Ajit Nair, Jivat~Neet Kaur, Ricardo Usbeck, and Chris Biemann.
\newblock Modern baselines for sparql semantic parsing.
\newblock In \emph{Proceedings of the 45th International ACM SIGIR Conference on Research and Development in Information Retrieval}, SIGIR ’22. ACM, July 2022.
\newblock \doi{10.1145/3477495.3531841}.
\newblock URL \url{http://dx.doi.org/10.1145/3477495.3531841}.

\bibitem[Kudo and Richardson(2018)]{kudo-richardson-2018-sentencepiece}
Taku Kudo and John Richardson.
\newblock {S}entence{P}iece: A simple and language independent subword tokenizer and detokenizer for neural text processing.
\newblock In Eduardo Blanco and Wei Lu, editors, \emph{Proceedings of the 2018 Conference on Empirical Methods in Natural Language Processing: System Demonstrations}, pages 66--71, Brussels, Belgium, November 2018. Association for Computational Linguistics.
\newblock \doi{10.18653/v1/D18-2012}.
\newblock URL \url{https://aclanthology.org/D18-2012}.

\bibitem[Dubey et~al.(2019)Dubey, Banerjee, Abdelkawi, and Lehmann]{lcquad2}
Mohnish Dubey, Debayan Banerjee, Abdelrahman Abdelkawi, and Jens Lehmann.
\newblock Lc-quad 2.0: A large dataset for complex question answering over wikidata and dbpedia.
\newblock In Chiara Ghidini, Olaf Hartig, Maria Maleshkova, Vojt{\v{e}}ch Sv{\'a}tek, Isabel Cruz, Aidan Hogan, Jie Song, Maxime Lefran{\c{c}}ois, and Fabien Gandon, editors, \emph{The Semantic Web -- ISWC 2019}, pages 69--78, Cham, 2019. Springer International Publishing.
\newblock ISBN 978-3-030-30796-7.

\bibitem[Perevalov et~al.(2022{\natexlab{b}})Perevalov, Diefenbach, Usbeck, and Both]{perevalov2022qald}
Aleksandr Perevalov, Dennis Diefenbach, Ricardo Usbeck, and Andreas Both.
\newblock Qald-9-plus: A multilingual dataset for question answering over dbpedia and wikidata translated by native speakers.
\newblock In \emph{2022 IEEE 16th International Conference on Semantic Computing (ICSC)}, pages 229--234. IEEE, 2022{\natexlab{b}}.

\bibitem[Usbeck et~al.(2019)Usbeck, R{\"o}der, Hoffmann, Conrad, Huthmann, Ngonga-Ngomo, Demmler, and Unger]{gerbil-qa}
Ricardo Usbeck, Michael R{\"o}der, Michael Hoffmann, Felix Conrad, Jonathan Huthmann, Axel-Cyrille Ngonga-Ngomo, Christian Demmler, and Christina Unger.
\newblock {Benchmarking Question Answering Systems}.
\newblock \emph{Semantic Web}, 10\penalty0 (2):\penalty0 293--304, January 2019.
\newblock \doi{10.3233/SW-180312}.
\newblock URL \url{http://www.semantic-web-journal.net/system/files/swj1578.pdf}.

\bibitem[Rajbhandari et~al.(2020)Rajbhandari, Rasley, Ruwase, and He]{rajbhandari2020zero}
Samyam Rajbhandari, Jeff Rasley, Olatunji Ruwase, and Yuxiong He.
\newblock Zero: Memory optimizations toward training trillion parameter models, 2020.

\bibitem[Shivashankar et~al.(2022)Shivashankar, Benmaarouf, and Steinmetz]{Shivashankar2022FromGraph}
K.~Shivashankar, K.~Benmaarouf, and N.~Steinmetz.
\newblock From graph to graph: Amr to sparql.
\newblock In \emph{Proceedings of the 7th Natural Language Interfaces for the Web of Data (NLIWoD) co-located with the 19th European Semantic Web Conference (ESWC 2022)}, 2022.

\bibitem[Baramiia et~al.(2022)Baramiia, Rogulina, Petrakov, Kornilov, and Razzhigaev]{Baramiia2022RankingApproach}
N.~Baramiia, A.~Rogulina, S.~Petrakov, V.~Kornilov, and A.~Razzhigaev.
\newblock Ranking approach to monolingual question answering over knowledge graphs.
\newblock In \emph{Proceedings of the 7th Natural Language Interfaces for the Web of Data (NLIWoD) co-located with the 19th European Semantic Web Conference (ESWC 2022)}, 2022.

\bibitem[Jurafsky and Martin(2024)]{JurafskyMartin2024}
Daniel Jurafsky and James~H. Martin.
\newblock Speech and language processing (3rd edition draft).
\newblock \url{https://web.stanford.edu/~jurafsky/slp3/}, 2024.
\newblock Accessed on: 15.02.2024.

\end{thebibliography}

\end{document}